\documentclass[runningheads]{llncs}
\usepackage{graphicx}
\usepackage{amsmath,amssymb} 
\usepackage{color}
\usepackage[width=122mm,left=12mm,paperwidth=146mm,height=193mm,top=12mm,paperheight=217mm]{geometry}
\usepackage{soul}
\usepackage{hyperref}

\begin{document}
\title{Learning to Forecast and Refine Residual Motion for Image-to-Video Generation}

\titlerunning{Learning to Forecast and Refine Residual Motion}
%
\author{Long~Zhao\inst{1} \and
Xi~Peng\inst{2} \and
Yu~Tian\inst{1} \and
Mubbasir~Kapadia\inst{1} \and
Dimitris~Metaxas\inst{1}}
%
\authorrunning{L.~Zhao et al.}
%

\institute{Rutgers University, Piscataway, USA\\
\email{\{lz311,yt219,mk1353,dnm\}@cs.rutgers.edu} \and
Binghamton University, Binghamton, USA\\
\email{xpeng@binghamton.edu}}
\maketitle              
\begin{abstract}
We consider the problem of image-to-video translation, where an input image is translated into an output video containing motions of a single object. Recent methods for such problems typically train transformation networks to generate future frames conditioned on the structure sequence. Parallel work has shown that short high-quality motions can be generated by spatiotemporal generative networks that leverage temporal knowledge from the training data. We combine the benefits of both approaches and propose a two-stage generation framework where videos are generated from structures and then refined by temporal signals. To model motions more efficiently, we train networks to learn residual motion between the current and future frames, which avoids learning motion-irrelevant details. We conduct extensive experiments on two image-to-video translation tasks: facial expression retargeting and human pose forecasting. Superior results over the state-of-the-art methods on both tasks demonstrate the effectiveness of our approach.\footnote{The project website is publicly available at \url{https://garyzhao.github.io/FRGAN}.}

\keywords{Video generation \and Motion forecasting \and Residual learning}
\end{abstract}
\newcommand{\p}[1]{\paragraph{\textbf{\textup{#1}}}}

\section{Introduction}

Recently, Generative Adversarial Networks (GANs)~\cite{goodfellow2014generative} have attracted a lot of research interests, as they can be utilized to synthesize realistic-looking images or videos for various vision applications~\cite{lu2017safetynet,ma2017pose,villegas2017learning,zhang2017image,zhang2018photographic,zhang2018translating}. Compared with image generation, synthesizing videos is more challenging, since the networks need to learn the appearance of objects as well as their motion models. In this paper, we study a form of classic problems in video generation that can be framed as image-to-video translation tasks, where a system receives one or more images as the input and translates it into a video containing realistic motions of a \textit{single} object. Examples include facial expression retargeting~\cite{laine2017production,olszewski2017realistic,thies2016face}, future prediction~\cite{tulyakov2017mocogan,villegas2017decomposing,vondrick2016generating}, and human pose forecasting~\cite{chao2017forecasting,fragkiadaki2015recurrent,villegas2017learning}.

One approach for the long-term future generation~\cite{villegas2017learning,xingjian2015convolutional} is to train a transformation network that translates the input image into each future frame separately conditioned by a sequence of structures. It suggests that it is beneficial to incorporate high-level structures during the generative process. In parallel, recent work~\cite{ji20133d,simonyan2014two,tran2015learning,tulyakov2017mocogan,vondrick2016generating} has shown that temporal visual features are important for video modeling. Such an approach produces temporally coherent motions with the help of spatiotemporal generative networks but is poor at long-term conditional motion generation since no high-level guidance is provided.

\begin{figure}[t]
\centering
\includegraphics[width=\linewidth]{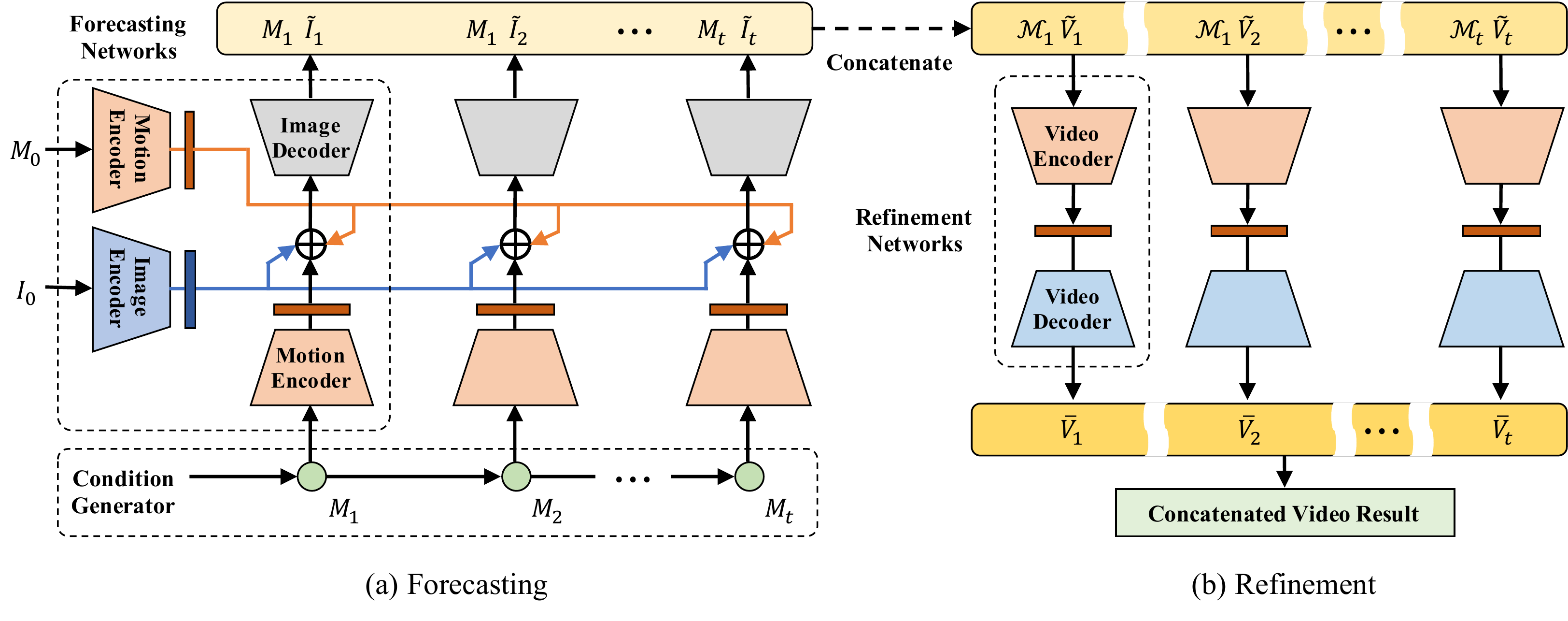}
\vspace{-18pt}
\caption{Method overview. Videos are (a) generated from conditions and then (b) refined. Our framework consists of three components: a condition generator, motion forecasting networks and refinement networks. Each part is explained in the corresponding section.}
\label{fig:overview}
\end{figure}

In this paper, we combine the benefits of these two methods. Our approach includes two motion transformation networks as shown in Figure~\ref{fig:overview}, where the entire video is synthesized in a generation and then refinement manner. In the generation stage, the \textit{motion forecasting networks} observe a single frame from the input and generate all future frames individually, which are conditioned by the structure sequence predicted by a \textit{motion condition generator}. This stage aims to generate a coarse video where the spatial structures of the motions are preserved. In the refinement stage, spatiotemporal \textit{motion refinement networks} are used for refining the output from the previous stage. It performs the generation guided by temporal signals, which targets at producing temporally coherent motions.

For more effective motion modeling, two transformation networks are trained in the \textit{residual space}. Rather than learning the mapping from the structural conditions to motions directly, we force the networks to learn the differences between motions occurring in the current and future frames. The intuition is that learning only the residual motion avoids the redundant motion-irrelevant information, such as static backgrounds, which remains unchanged during the transformation. Moreover, we introduce a novel network architecture using \textit{dense connections} for decoders. It encourages reusing spatially different features and thus yields realistic-looking results.

We experiment on two tasks: facial expression retargeting and human pose forecasting. Success in either task requires reasoning realistic spatial structures as well as temporal semantics of the motions. Strong performances on both tasks demonstrate the effectiveness of our approach. In summary, our work makes the following \textit{contributions}:

\begin{itemize}
\item We devise a novel two-stage generation framework for image-to-video translation, where the future frames are generated according to the spatial structure sequence and then refined with temporal signals;
\item We investigate learning residual motion for video generation, which focuses on the motion-specific knowledge and avoids learning redundant or irrelevant details from the inputs;
\item Dense connections between layers of decoders are introduced to encourage spatially different feature reuse during the generation process, which yields more realistic-looking results;
\item We conduct extensive experimental validation on standard datasets which both quantitatively and subjectively compares our method with the state-of-the-arts to demonstrate the effectiveness of the proposed algorithm.
\end{itemize}

\section{Related Work}
\label{sect:relatedwork}

Deep learning techniques have improved the accuracy of various vision systems~\cite{peng2016recurrent,peng2015circle,tang2018quantized,tang2018cu}. Especially, a lot of generative problems~\cite{gulrajani2017improved,peng2018jointly,perarnau2016invertible,tian2018cr} have been solved by GANs~\cite{goodfellow2014generative}. However, traditional frameworks fail to handle complicated tasks, e.g., to generate fine-grained images or videos with large motion changes. Recent approaches~\cite{ma2017pose,wei2018learning,han2017stackgan} prove that coarse-to-fine strategy can handle these cases. Our model also employs this strategy for video generation.

Xiong et al.~\cite{wei2018learning} proposed an algorithm to generate video in two stages, but there are important differences between their work and ours. First, \cite{wei2018learning} is proposed for time-lapse videos while we can generate general videos. Second, we make use of structure conditions to guide the generation in the first stage but \cite{wei2018learning} models this stage with 3D convolutional networks. Finally, we can make long-term predictions while \cite{wei2018learning} only generates videos with fixed length.

\p{Video Generation.} Recent methods~\cite{mathieu2016deep,villegas2017decomposing,villegas2017learning,xingjian2015convolutional} solve image-to-video generation problem by training transformation networks that translate the input image into each future frame separately, together with a generator predicting the structure sequence which conditions the future frames. However, due to the absence of pixel-level temporal knowledge during the training process, motion artifacts can be observed from the results of these methods.

Other approaches explore learning temporal visual features from video with spatiotemporal networks. Ji et al.~\cite{ji20133d} showed how 3D convolutional networks could be applied to human action recognition. Tran et al.~\cite{tran2015learning} employed spatiotemporal 3D convolutions to model features encoded in videos. Vondrick et al.~\cite{vondrick2016generating} built a model to generate scene dynamics with 3D generative adversarial networks. Our method differs from the two-stream model of~\cite{vondrick2016generating} in two aspects. First, our residual motion map disentangles motion from the input: the generated frame is conditioned on the current and future motion structures. Second, we can control object motions in future frames efficiently by using structure conditions. Therefore, our method can be applied to motion manipulation problems.

\p{Dense Connections.} Recent studies~\cite{huang2018condensenet,huang2017densely} used dense connections for image classification. They have proven that dense connections for encoders strengthen feature propagation and also encourage feature reuse. Instead, we introduce dense connections for decoders. Compared with multi-scale feature fusion in~\cite{liu2017voxelflow} where feature maps are only concatenated to the last layer of the network, our dense connections upsample and concatenate feature maps with different scales to all intermediate layers. Our approach is more efficient at feature re-use when utilized for generation, which yields sharper and more realistic-looking results.

\section{Method}
\label{sect:method}

As shown in Figure~\ref{fig:overview}, our framework consists of three components: a motion condition generator $G_{C}$, an image-to-image transformation network $G_{M}$ for forecasting motion conditioned by $G_{C}$ to each future frame individually, and a video-to-video transformation network $G_{R}$ which aims to refine the video clips concatenated from the output of $G_{M}$. $G_{C}$ is a task-specific generator that produces a sequence of structures to condition the motion of each future frame. Two discriminators are utilized for adversarial learning, where $D_I$ differentiates real frames from generated ones and $D_V$ is employed for video clips. In the following sections, we explain how each component is designed respectively.

\subsection{Motion Condition Generators}
\label{sect:method:condition}

In this section, we illustrate how the motion condition generators $G_{C}$ are implemented for two image-to-video translation tasks: facial expression retargeting and human pose forecasting. One superiority of $G_{C}$ is that domain-specific knowledge can be leveraged to help the prediction of motion structures.

\begin{figure}[t]
\centering
\includegraphics[width=\linewidth]{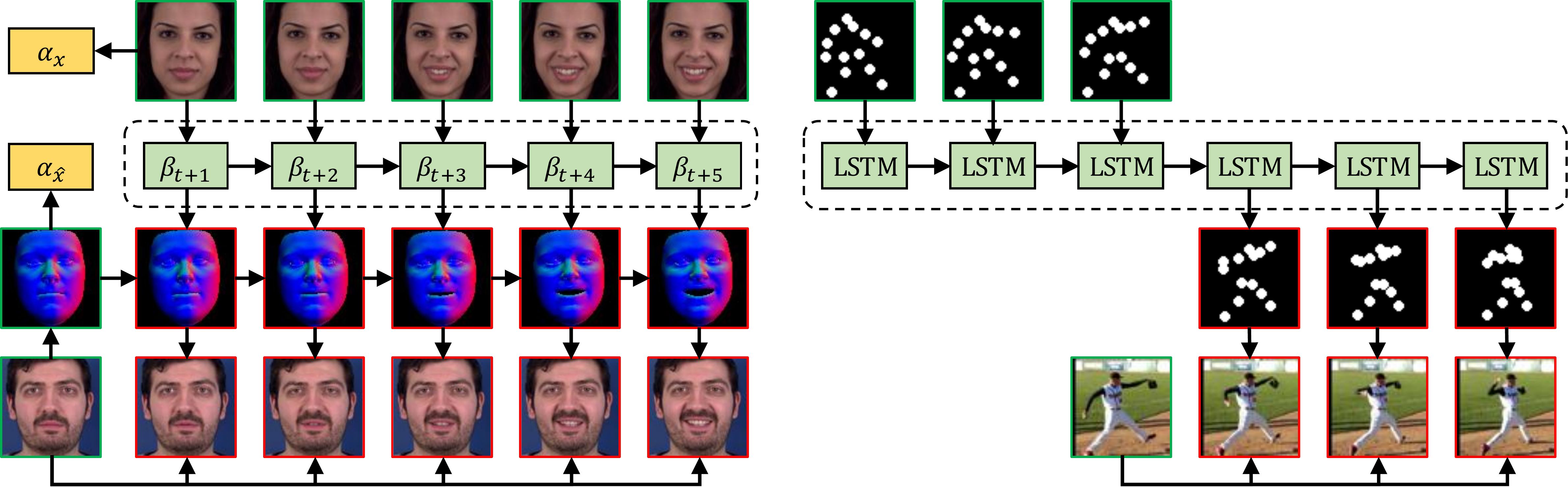}
\caption{Illustration of our motion condition generators designed for two tasks. \textit{Left}: For facial expression retargeting, 3D Morphable Model~\cite{blanz2003face} is utilized as domain knowledge to produce expression conditional sequence. \textit{Right}: For human pose forecasting, the pose is represented by the 2D positions of joints. The LSTM~\cite{fragkiadaki2015recurrent} observes a sequence of human pose inputs and predicts the next several timesteps.}
\label{fig:GC}
\end{figure}

\p{Facial Expression Retargeting.} As shown in Figure~\ref{fig:GC}, we utilize 3D Morphable Model~(3DMM)~\cite{blanz2003face} to model the sequence of expression motions. Given a video containing expression changes of an actor $x$, it can be parameterized with $\alpha_{x}$ and $(\beta_{t}, \beta_{t+1}, \dots, \beta_{t+k})$ using 3DMM, where $\alpha_{x}$ represents the facial identity and $\beta_{t}$ is the expression coefficients in the frame $t$. In order to retarget the sequence of expressions to another actor $\hat{x}$, we compute the facial identity vector $\alpha_{\hat{x}}$ and combine it with $(\beta_{t}, \beta_{t+1}, \dots, \beta_{t+k})$ to reconstruct a new sequence of 3D face models with corresponding facial expressions. The conditional motion maps are the normal maps calculated from the 3D models respectively.

\p{Human Pose Forecasting.} We follow~\cite{villegas2017learning} to implement an LSTM architecture~\cite{fragkiadaki2015recurrent} as the human pose predictor. The human pose of each frame is represented by the 2D coordinate positions of joints. The LSTM observes consecutive pose inputs to identify the type of motion, and then predicts the pose for the next period of time. An example is shown in Figure~\ref{fig:GC}. Note that the motion map is calculated by mapping the output 2D coordinates from the LSTM to heatmaps and concatenating them on depth.

\subsection{Motion Forecasting Networks}
\label{sect:method:forecasting}

\begin{figure}[t]
\centering
\includegraphics[width=\linewidth]{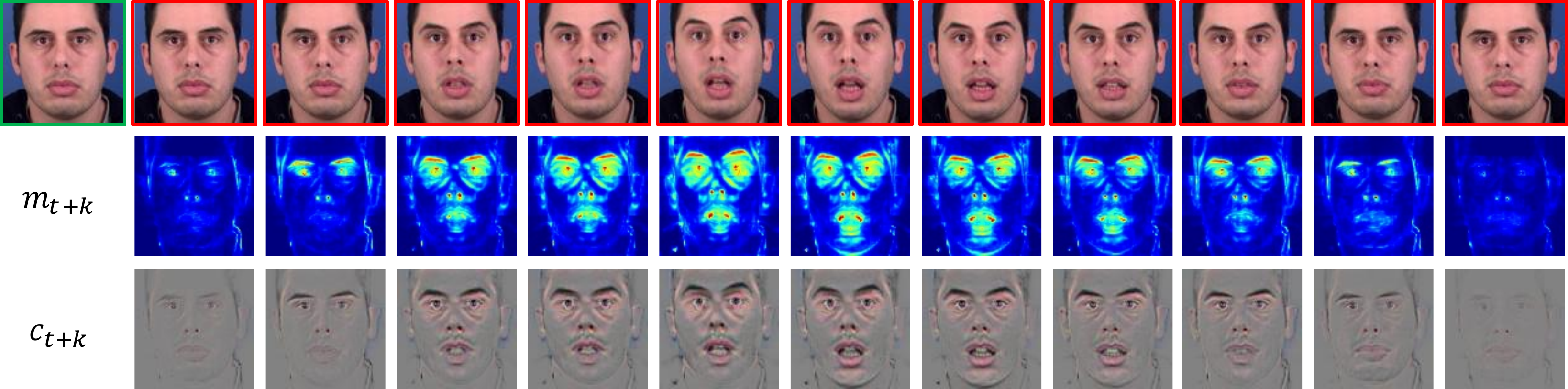}
\caption{Illustration of our residual formulation. We disentangle the motion differences between the input and future frames into a residual motion map $m_{t+k}$ and a residual content map $c_{t+k}$. Compared with the difference map directly computed from them, our formulation makes the learning task much easier.}
\label{fig:residual}
\end{figure}

Starting from the frame $I_{t}$ at time $t$, our network synthesizes the future frame $I_{t+k}$ by predicting the residual motion between them. Previous work~\cite{ma2017pose,shen2017learning} implemented this idea by letting the network estimate a difference map between the input and output, which can be denoted as:
\begin{equation}
\label{eq:GM:simRes}
I_{t+k} = I_{t} + G_M(I_{t}|M_{t}, M_{t+k}),
\end{equation}
where $M_{t}$ is the motion map which conditions $I_{t}$. However, this straightforward formulation easily fails when employed to handle videos including large motions, since learning to generate a combination of residual changes from both dynamic and static contents in a single map is quite difficult. Therefore, we introduce an enhanced formulation where the transformation is disentangled into a residual motion map $m_{t+k}$ and a residual content map $c_{t+k}$ with the following definition:
\begin{equation}
\label{eq:GM:disRes}
I_{t+k} = \underbrace{m_{t+k} \odot c_{t+k}}_{\text{residual motion}} + \underbrace{(1 - m_{t+k}) \odot I_{t}}_{\text{static content}},
\end{equation}
where both $m_{t+k}$ and $c_{t+k}$ are predicted by $G_M$, and $\odot$ is element-wise multiplication. Intuitively, $m_{t+k} \in [0, 1]$ can be viewed as a spatial mask that highlights where the motion occurs. $c_{t+k}$ is the content of the residual motions. By summing the residual motion with the static content, we can obtain the final result. Note that as visualized in Figure~\ref{fig:residual}, $m_{t+k}$ forces $G_M$ to reuse the static part from the input and concentrate on inferring dynamic motions.

\begin{figure}[t]
\centering
\includegraphics[width=\linewidth]{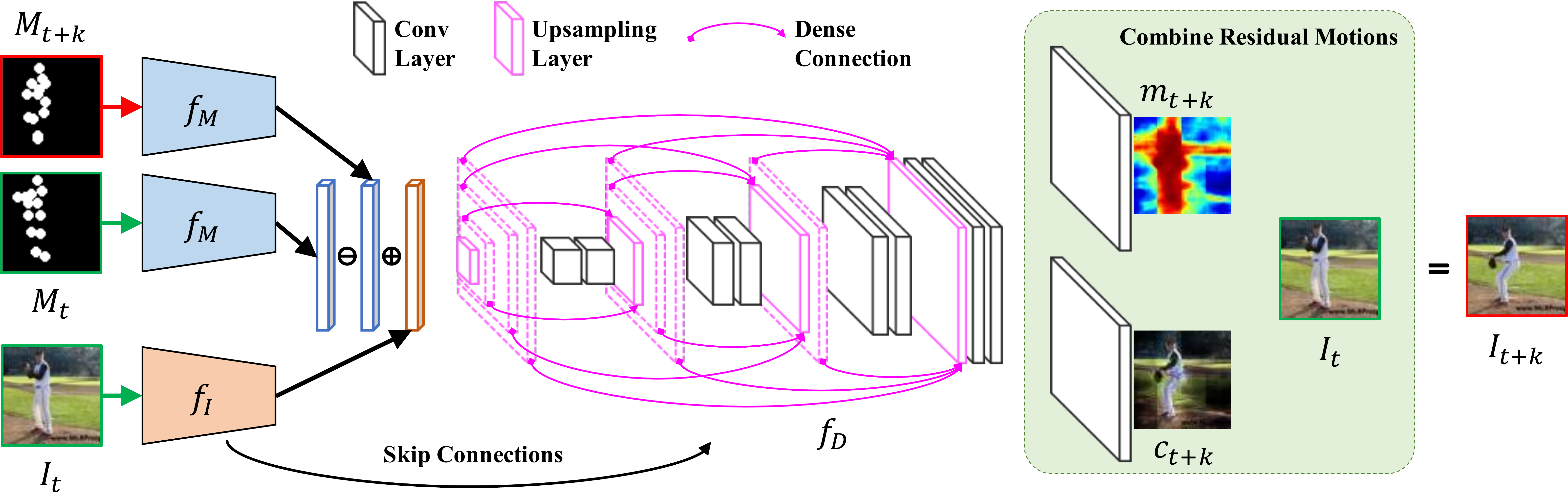}
\caption{Architecture of our motion forecasting network $G_M$. The network observes the input frame $I_t$ with its corresponding motion map $M_t$, and the motion map of the future frame $M_{t+k}$. Through analogy learning, the network estimates the \textit{residual motion} between the current frame $I_t$ and future frame $I_{t+k}$. Note that the \textit{dashed} layers upsample the inputs and connect them to the subsequent dense blocks.}
\label{fig:GM}
\end{figure}

\p{Architecture.} Figure~\ref{fig:GM} shows the architecture of $G_M$, which is inspired by the visual-structure analogy learning~\cite{reed2015deep}. The future frame $I_{t+k}$ can be generated by transferring the structure differences from $M_t$ to $M_{t+k}$ to the input frame $I_t$. We use a motion encoder $f_M$, an image encoder $f_I$ and a residual content decoder $f_D$ to model this concept. And the residual motion is learned by:
\begin{equation}
\label{eq:GM:analogy}
\Delta(I_{t+k}, I_t) = f_D(f_M(M_{t+k}) - f_M(M_{t}) + f_I(I_t)).
\end{equation}
Intuitively, $f_M$ aims to identify key motion features from the motion map containing high-level structural information; $f_I$ learns to map the appearance model of the input into an embedding space, where the motion feature transformations can be easily imposed to generate the residual motion; $f_D$ learns to decode the embedding. Note that we add skip connections~\cite{ronneberger2015unet} between $f_I$ and $f_D$, which makes it easier for $f_D$ to reuse features of static objects learned from $f_I$.

Huang et al.~\cite{huang2018condensenet,huang2017densely} introduce dense connections to enhance feature propagation and reuse in the network. We argue that this is an appealing property for motion transformation networks as well, since in most cases the output frame shares similar high-level structure with the input frame. Especially, dense connections make it easy for the network to reuse features of different spatial positions when large motions are involved in the image. The decoder of our network thus consists of multiple \textit{dense connections}, each of which connects different dense blocks. A dense block contains two $3 \times 3$ convolutional layers. The output of a dense block is connected to the first convolutional layers located in \textit{all} subsequent blocks in the network. As dense blocks have different feature resolutions, we upsample feature maps with lower resolutions when we use them as inputs into higher resolution layers.

\p{Training Details.} Given a video clip, we train our network to perform random jumps in time to learn forecasting motion changes. To be specific, for every iteration at training time, we sample a frame $I_t$ and its corresponding motion map $M_t$ given by $G_C$ at time $t$, and then force it to generate frame $I_{t+k}$ given motion map $M_{t+k}$. Note that in order to let our network perform learning in the entire residual motion space, $k$ is also randomly defined for each iteration. On the other hand, learning with jumps in time can prevent the network from falling into suboptimal parameters as well~\cite{villegas2017learning}. Our network is trained to minimize the following objective function:
\begin{equation}
\label{eq:GM:lossGM}
\mathcal{L}_{G_M} = \mathcal{L}_{rec}(I_{t+k}, \tilde{I}_{t+k}) + \mathcal{L}_{r}(m_{t+k}) + \mathcal{L}_{gen}.
\end{equation}

$\mathcal{L}_{rec}$ is the reconstruction loss defined in the image space which measures the pixel-wise differences between the predicted and target frames:
\begin{equation}
\label{eq:GM:lossRec}
\mathcal{L}_{rec}(I_{t+k}, \tilde{I}_{t+k}) = {\| I_{t+k} - \tilde{I}_{t+k} \|}_1,
\end{equation}
where $\tilde{I}_{t+k}$ denotes the frame predicted by $G_M$. The reconstruction loss intuitively offers guidance for our network in making a rough prediction that preserves most content information of the target image. More importantly, it leads the result to share similar structure information with the input image. $\mathcal{L}_{r}$ is an L-1 norm regularization term defined as:
\begin{equation}
\label{eq:GM:lossM}
\mathcal{L}_{r}(m_{t+k}) = {\| m_{t+k} \|}_1,
\end{equation}
where $m_{t+k}$ is the residual motion map predicted by $G_M$. It forces the predicted motion changes to be sparse, since dynamic motions always occur in local positions of each frame while the static parts (e.g., background objects) should be unchanged. $\mathcal{L}_{gen}$ is the adversarial loss that enables our model to generate realistic frames and reduce blurs, and it is defined as:
\begin{equation}
\label{eq:GM:lossGen}
\mathcal{L}_{gen} = -D_I([\tilde{I}_{t+k}, M_{t+k}]),
\end{equation}
where $D_I$ is the discriminator for images in adversarial learning. We concatenate the output of $G_M$ and motion map $M_{t+k}$ as the input of $D_I$ and make the discriminator conditioned on the motion~\cite{mirza2014conditional}.

Note that we follow WGAN~\cite{arjovsky2017wasserstein,gulrajani2017improved} to train $D_I$ to measure the Wasserstein distance between distributions of the real images and results generated from $G_M$. During the optimization of $D_I$, the following loss function is minimized:
\begin{equation}
\label{eq:GM:lossDI}
\mathcal{L}_{D_I} = D_I([\tilde{I}_{t+k}, M_{t+k}]) - D_I([I_{t+k}, M_{t+k}]) + \lambda \cdot \mathcal{L}_{gp},
\end{equation}
\begin{equation}
\label{eq:GM:lossGP}
\mathcal{L}_{gp} = ({\| \nabla_{[\hat{I}_{t+k}, M_{t+k}]} D_I([\hat{I}_{t+k}, M_{t+k}]) \|}_2 - 1)^2,
\end{equation}
where $\lambda$ is experimentally set to 10. $\mathcal{L}_{gp}$ is the gradient penalty term proposed by~\cite{gulrajani2017improved} where $\hat{I}_{t+k}$ is sampled from the interpolation of $I_{t+k}$ and $\tilde{I}_{t+k}$, and we extend it to be conditioned on the motion $M_{t+k}$ as well. The adversarial loss in combination with the rest of loss terms allows our network to generate high-quality frames given the motion conditions.

\subsection{Motion Refinement Networks}
\label{sect:method:refinement}

Let $\tilde{V}_{t} = [\tilde{I}_{t+1}, \tilde{I}_{t+2}, \dots, \tilde{I}_{t+K}]$ be the video clip with length $K$ temporally concatenated from the outputs of $G_M$. The goal of the motion refinement network $G_R$ is to refine $\tilde{V}_{t}$ to be more temporally coherent, which is achieved by performing pixel-level refinement with the help of spatiotemporal generative networks. We extends Equation~\ref{eq:GM:disRes} by adding one additional temporal dimension to let $G_R$ estimate the residual between the real video clip $V_{t}$ and $\tilde{V}_{t}$, which is defined as:
\begin{equation}
\label{eq:GR:res}
V_{t} = m_{t} \odot c_{t} + (1 - m_{t}) \odot \tilde{V}_{t},
\end{equation}
where $m_{t}$ is a spatiotemporal mask which selects either to be refined for each pixel location and timestep, while $c_{t}$ produces a spatiotemporal cuboid which stands for the refined motion content masked by $m_{t}$.

\begin{figure}[t]
\centering
\includegraphics[width=\linewidth]{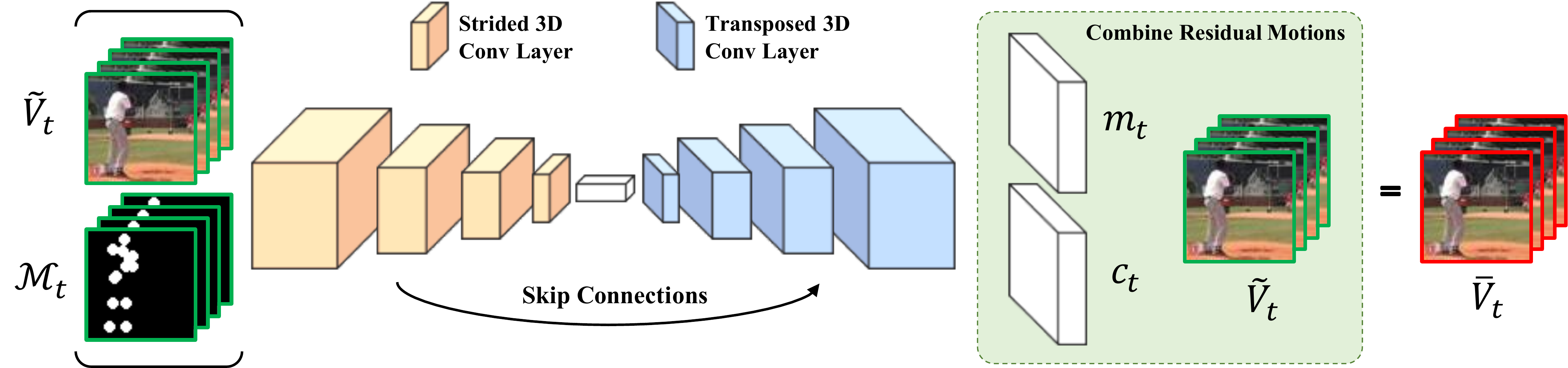}
\caption{Architecture of our motion refinement network $G_R$. The network receives temporally concatenated frames generated by $G_M$ together with their corresponding conditional motion map as the input and aims to refine the video clip to be more temporally coherent. It performs learning in the residual motion space as well.}
\label{fig:GR}
\end{figure}

\p{Architecture.} Our motion refinement network roughly follows the architectural guidelines of~\cite{vondrick2016generating}. As shown in Figure~\ref{fig:GR}, we do not use pooling layers, instead strided and fractionally strided convolutions are utilized for in-network downsampling and upsampling. We also add skip connections to encourage feature reuse. Note that we concatenate the frames with their corresponding conditional motion maps as the inputs to guide the refinement.

\p{Training Details.} The key requirement for $G_R$ is that the refined video should be temporal coherent in motion while preserving the annotation information from the input. To this end, we propose to train the network by minimizing a combination of three losses which is similar to Equation~\ref{eq:GM:lossGM}:
\begin{equation}
\label{eq:GM:lossGR}
\mathcal{L}_{G_R} = \mathcal{L}_{rec}(V_{t}, \bar{V}_{t}) + \mathcal{L}_{r}(m_{t}) + \bar{\mathcal{L}}_{gen},
\end{equation}
where $\bar{V}_{t}$ is the output of $G_R$. $\mathcal{L}_{rec}$ and $\mathcal{L}_{r}$ share the same definition with Equation~\ref{eq:GM:lossRec} and~\ref{eq:GM:lossM} respectively. $\mathcal{L}_{rec}$ is the reconstruction loss that aims at refining the synthesized video towards the ground truth with minimal error. Compared with the self-regularization loss proposed by~\cite{shrivastava2017learning}, we argue that the sparse regularization term $\mathcal{L}_{r}$ is also efficient to preserve the annotation information (e.g., the facial identity and the type of pose) during the refinement, since it force the network to only modify the essential pixels. $\bar{\mathcal{L}}_{gen}$ is the adversarial loss:
\begin{equation}
\label{eq:GR:lossGen}
\bar{\mathcal{L}}_{gen} = -D_V([\bar{V}_{t}, \mathcal{M}_{t}]) - \frac{1}{K}\sum_{i=1}^K D_I([\bar{I}_{t+i}, M_{t+i}]),
\end{equation}
where $\mathcal{M}_{t} = [M_{t+1}, M_{t+2}, \dots, M_{t+K}]$ is the temporally concatenated condition motion maps, and $\bar{I}_{t+i}$ is the $i$-th frame of $\bar{V}_{t}$. In the adversarial learning term $\bar{\mathcal{L}}_{gen}$, both $D_I$ and $D_V$ play the role to judge whether the input is a real video clip or not, providing criticisms to $G_R$. The image discriminator $D_I$ criticizes $G_R$ based on individual frames, which is trained to determine if each frame is sampled from a real video clip. At the same time, $D_V$ provides criticisms to $G_R$ based on the whole video clip, which takes a fixed length video clip as the input and judges if a video clip is sampled from a real video as well as evaluates the motions contained. As suggested by~\cite{tulyakov2017mocogan}, although $D_V$ alone should be sufficient, $D_I$ significantly improves the convergence and the final results of $G_R$.

We follow the same strategy as introduced in Equation~\ref{eq:GM:lossDI} to optimize $D_I$. Note that in each iteration, one pair of real and generated frames is randomly sampled from $V_{t}$ and $\bar{V}_{t}$ to train $D_I$. On the other hand, training $D_V$ is also based on the WGAN framework, where we extend it to spatiotemporal inputs. Therefore, $D_V$ is optimized by minimizing the following loss function:
\begin{equation}
\label{eq:GR:lossDV}
\mathcal{L}_{D_V} = D_V([\bar{V}_{t}, \mathcal{M}_{t}]) - D_V([V_t, \mathcal{M}_{t}]) + \lambda \cdot \mathcal{L}_{gp},
\end{equation}
\begin{equation}
\label{eq:GR:lossGP}
\mathcal{L}_{gp} = ({\| \nabla_{[\hat{V}_{t}, \mathcal{M}_{t}]} D_V([\hat{V}_{t}, \mathcal{M}_{t}]) \|}_2 - 1)^2,
\end{equation}
where $\hat{V}_{t}$ is sampled from the interpolation of $V_{t}$ and $\bar{V}_{t}$. Note that $G_R$, $D_I$ and $D_V$ are trained alternatively. To be specific, we update $D_I$ and $D_V$ in one step while fixing $G_R$; in the alternating step, we fix $D_I$ and $D_V$ while updating $G_R$.

\section{Experiments}
\label{sect:experiments}

We perform experiments on two image-to-video translation tasks: facial expression retargeting and human pose forecasting. For facial expression retargeting, we demonstrate that our method is able to combine domain-specific knowledge, such as 3DMM, to generate realistic-looking results. For human pose forecasting, experimental results show that our method yields high-quality videos when applied for video generation tasks containing complex motion changes.

\subsection{Settings and Databases}
\label{sect:experiments:settings}

To train our networks, we use Adam~\cite{kingma2014adam} for optimization with a learning rate of 0.0001 and momentums of 0.0 and 0.9. We first train the forecasting networks, and then train the refinement networks using the generated coarse frames. The batch size is set to 32 for all networks. Due to space constraints, we ask the reader to refer to the project website for the details of the network designs.

We use the \textit{MUG Facial Expression Database}~\cite{aifanti2010mug} to evaluate our approach on facial expression retargeting. This dataset is composed of 86 subjects (35 women and 51 men). We crop the face regions with regards to the landmark ground truth and scale them to $96 \times 96$. To train our networks, we use only the sequences representing one of the six facial expressions: anger, fear, disgust, happiness, sadness, and surprise. We evenly split the database into three groups according to the subjects. Two groups are used for training $G_M$ and $G_R$ respectively, and the results are evaluated on the last one. The 3D Basel Face Model~\cite{paysan2009bfm} serves as the morphable model to fit the facial identities and expressions for the condition generator $G_C$. We use~\cite{zhu2016face} to compute the 3DMM parameters for each frame. Note that we train $G_R$ to refine the video clips every 32 frames.

The \textit{Penn Action Dataset}~\cite{zhang2013from} consists of 2326 video sequences of 15 different human actions, which is used for evaluating our method on human pose forecasting. For each action sequence in the dataset, 13 human joint annotations are provided as the ground truth. To remove very noisy joint ground-truth in the dataset, we follow the setting of~\cite{villegas2017learning} to sub-sample the actions. Therefore, 8 actions including baseball pitch, baseball swing, clean and jerk, golf swing, jumping jacks, jump rope, tennis forehand, and tennis serve are used for training our networks. We crop video frames based on temporal tubes to remove as much background as possible while ensuring the human actions are in all frames, and then scale each cropped frame to $64 \times 64$. We evenly split the standard dataset into three sets. $G_M$ and $G_R$ are trained in the first two sets respectively, while we evaluate our models in the last set. We employ the same strategy as~\cite{villegas2017learning} to train the LSTM pose generator. It is trained to observe 10 inputs and predict 32 steps. Note that $G_R$ is trained to refine the video clips with the length of 16.

\subsection{Evaluation on Facial Expression Retargeting}
\label{sect:experiments:expression}

We compare our method to MCNet~\cite{villegas2017decomposing}, MoCoGAN~\cite{tulyakov2017mocogan} and Villegas~et~al.~\cite{villegas2017learning} on the MUG Database. For each facial expression, we randomly select one video as the reference, and retarget it to all the subjects in the testing set with different methods. Each method only observes the input frame of the target subject, and performs the generation based on it. Our method and~\cite{villegas2017learning} share the same 3DMM-based condition generator as introduced in Section~\ref{sect:method:condition}.

\p{Quantitative Comparison.} The quality of a generated video are measured by the Average Content Distance (ACD) as introduced in~\cite{tulyakov2017mocogan}. For each generated video, we make use of OpenFace~\cite{amos2016openface}, which outperforms human performance in the face recognition task, to measure the video quality. OpenFace produces a feature vector for each frame, and then the ACD is calculated by measuring the L-2 distance of these vectors. We introduce two variants of the ACD in this experiment. The ACD-I is the average distance between each generated frame and the original input frame. It aims to judge if the facial identity is well-preserved in the generated video. The ACD-C is the average pairwise distance of the per-frame feature vectors in the generated video. It measures the content consistency of the generated video.

\begin{figure}[t]
\centering
\includegraphics[width=\linewidth]{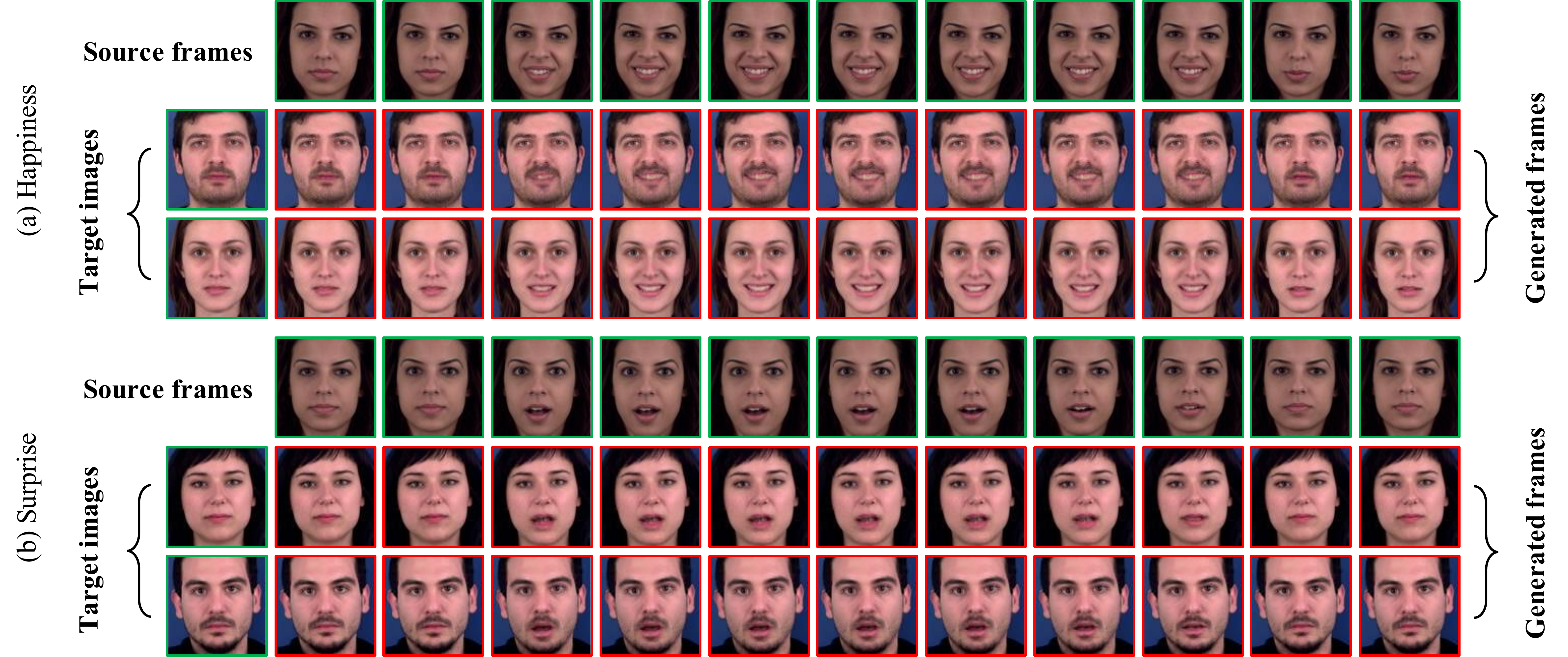}
\caption{Examples of facial expression retargeting using our algorithm on the MUG Database~\cite{aifanti2010mug}. We show two expressions as an illustration: (a) happiness and (b) surprise. The reference video and the input target images are highlighted in \textit{green}, while the generated frames are highlighted in \textit{red}. The results are sampled every 8 frames.}
\label{fig:MUG}
\end{figure}

\setlength{\tabcolsep}{2pt}
\begin{table}[t]
\begin{minipage}[t]{.44\linewidth}
\begin{center}
\caption{Video generation quality comparison on the MUG Dataset~\cite{aifanti2010mug}. We also compute the ACD-* score for the training set, which is the reference.}
\label{tbl:MUG:ACD}
\begin{tabular}{c|c|c}
\hline
Methods & ACD-I & ACD-C\\ \hline\hline
MCNet~\cite{villegas2017decomposing} & 0.545 & 0.322\\ \hline
Villegas et al.~\cite{villegas2017learning} & 0.683 & 0.130\\ \hline
MoCoGAN~\cite{tulyakov2017mocogan} & 0.291 & 0.205\\ \hline
Ours & \textbf{0.184} & \textbf{0.107}\\ \hline
Reference & 0.109 & 0.098\\ \hline
\end{tabular}
\end{center}
\end{minipage}
\hfill
\begin{minipage}[t]{.52\linewidth}
\begin{center}
\caption{Average user preference score (the average number of times, a user prefers our result to the competing one) on the MUG Dataset~\cite{aifanti2010mug}. Our results own higher preference scores compared with the others.}
\label{tbl:MUG:AUP}
\begin{tabular}{l|c}
\hline
Methods & Preference (\%) \\ \hline\hline
Ours / MCNet~\cite{villegas2017decomposing} & \textbf{84.2} / 15.8\\ \hline
Ours / Villegas et al.~\cite{villegas2017learning} & \textbf{74.6} / 25.4\\ \hline
Ours / MoCoGAN~\cite{tulyakov2017mocogan} & \textbf{62.5} / 37.5\\ \hline
\end{tabular}
\end{center}
\end{minipage}
\end{table}
\setlength{\tabcolsep}{1.4pt}

Table~\ref{tbl:MUG:ACD} summarizes the comparison results. From the table, we find that our method achieves ACD-* scores both lower than 0.2, which is substantially better than the baselines. One interesting observation is that~\cite{villegas2017learning} has the worst ACD-I but its ACD-C is the second best. We argue that this is due to the high-level information offered by our 3DMM-based condition generator, which plays a vital role for producing content consistency results. Our method outperforms other state-of-the-arts, since we utilize both domain knowledge (3DMM) and temporal signals for video generation. We show that it is greatly beneficial to incorporate both factors into the generative process.

We also conduct a user study to quantitatively compare these methods. For each method, we randomly select 10 videos for each expression. We then randomly pair the videos generated by ours with the videos from one of the competing methods to form 54 questions. For each question, 3 users are asked to select the video which is more realistic. To be fair, the videos from different methods are shown in random orders. We report the average user preference scores (the average number of times, a user prefers our result to the competing one) in Table~\ref{tbl:MUG:AUP}. We find that the users consider the videos generated by ours more realistic most of the time. This is consistent with the ACD results in Table~\ref{tbl:MUG:ACD}, in which our method substantially outperforms the baselines.

\p{Visual Results.} In Figure~\ref{fig:MUG}, we show the visual results (the expressions of happiness and surprise) generated by our method. We observe that our method is able to generate realistic motions while the facial identities are well-preserved. We hypothesize that the domain knowledge (3DMM) employed serves as a good prior which improves the generation. More visual results of different expressions and subjects are given on the project website.

\subsection{Evaluation on Human Pose Forecasting}
\label{sect:experiments:pose}

We compare our approach with VGAN~\cite{vondrick2016generating}, Mathieu~et~al.~\cite{mathieu2016deep} and Villegas~et~al.~\cite{villegas2017learning} on the Penn Action Dataset. We produce the results of their models according to their papers or reference codes. For fair comparison, we generate videos with 32 generated frames using each method, and evaluate them starting from the first frame. Note that we train an individual VGAN for different action categories with randomly picked video clips from the dataset, while one network among all categories are trained for every other method. Both \cite{villegas2017learning} and our method perform the generation based on the pre-trained LSTM provided by~\cite{villegas2017learning}, and we train \cite{villegas2017learning} through the same strategy of our motion forecasting network $G_M$.

\p{Implementation.} Following the settings of~\cite{villegas2017learning}, we engage the feature similarity loss term $\mathcal{L}_{feat}$ for our motion forecasting network $G_M$ to capture the appearance ($C_1$) and structure ($C_2$) of the human action. This loss term is added to Equation~\ref{eq:GM:lossGM}, which is defined as:
\begin{equation}
\label{eq:GM:lossFeat}
\mathcal{L}_{feat} = {\| C_1(I_{t+k}) - C_1(\tilde{I}_{t+k}) \|}_2^2 + {\| C_2(I_{t+k}) - C_2(\tilde{I}_{t+k}) \|}_2^2,
\end{equation}
where we use the last convolutional layer of the VGG16 Network~\cite{simonyan2015vgg} as $C_1$, and the last layer of the Hourglass Network~\cite{newell2016stacked} as $C_2$. Note that we compute the bounding box according to the group truth to crop the human of interest for each frame, and then scale it to $224 \times 224$ as the input of the VGG16.

\begin{figure}[t]
\centering
\includegraphics[width=0.24\linewidth]{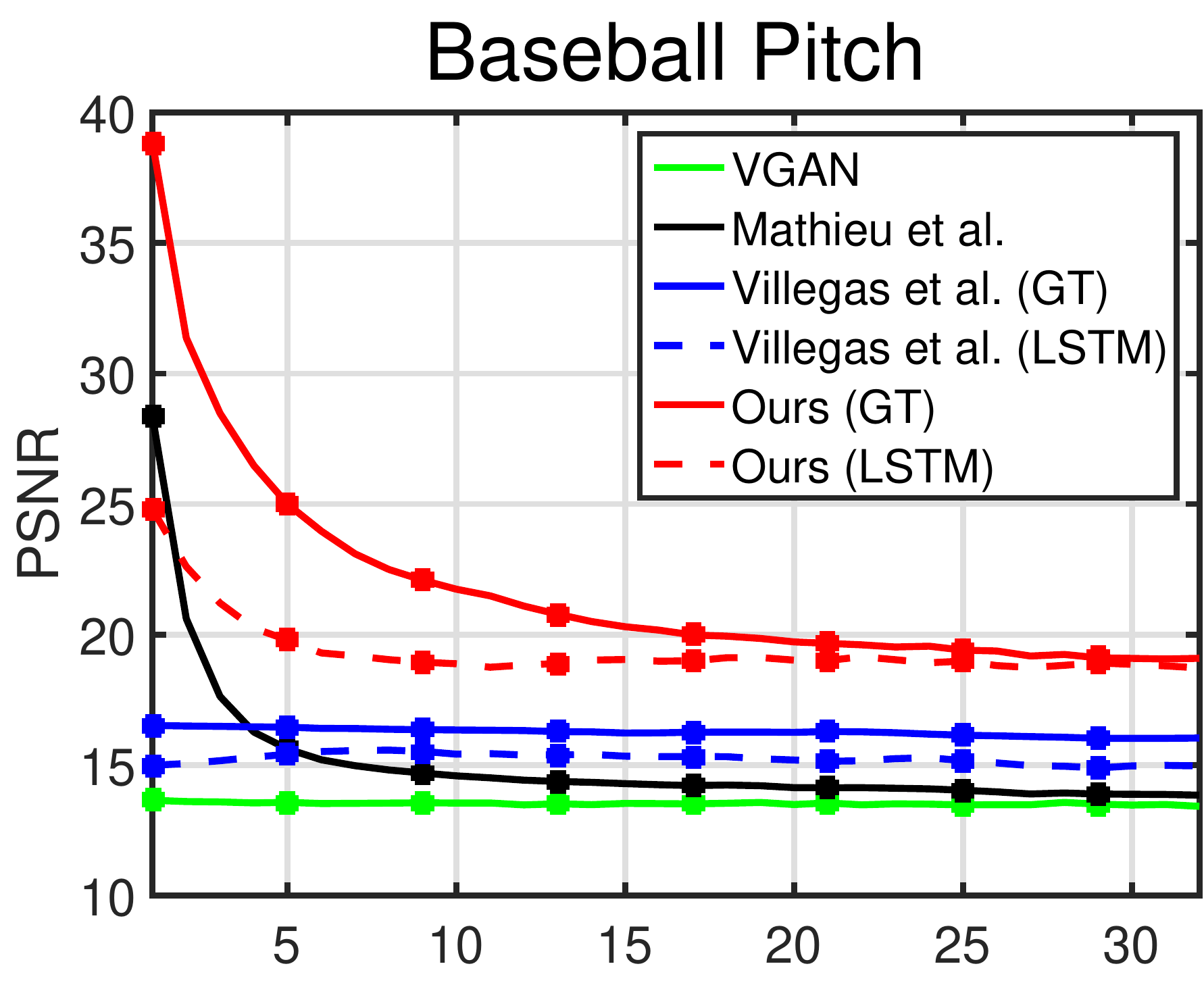}
\includegraphics[width=0.24\linewidth]{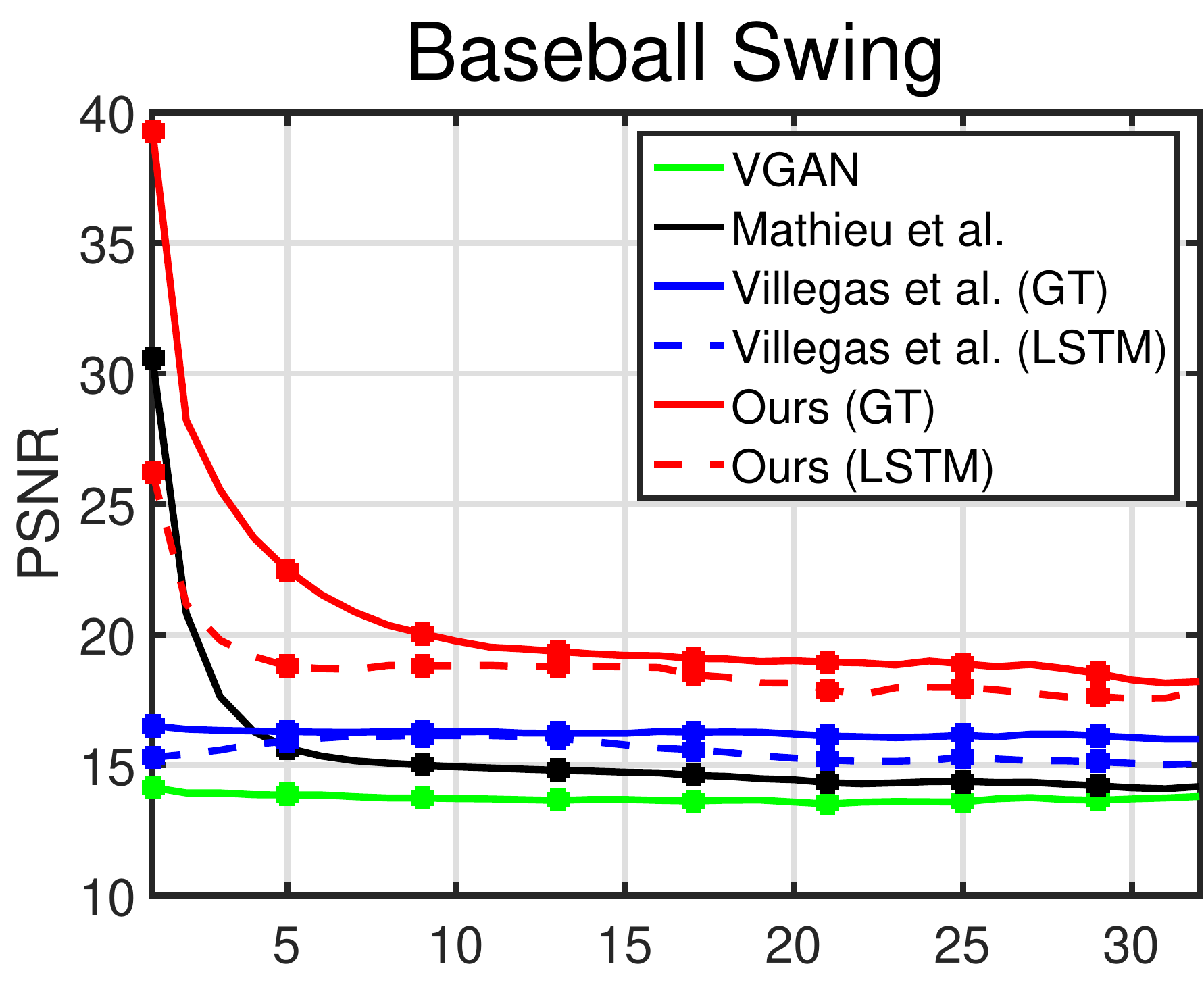}
\includegraphics[width=0.24\linewidth]{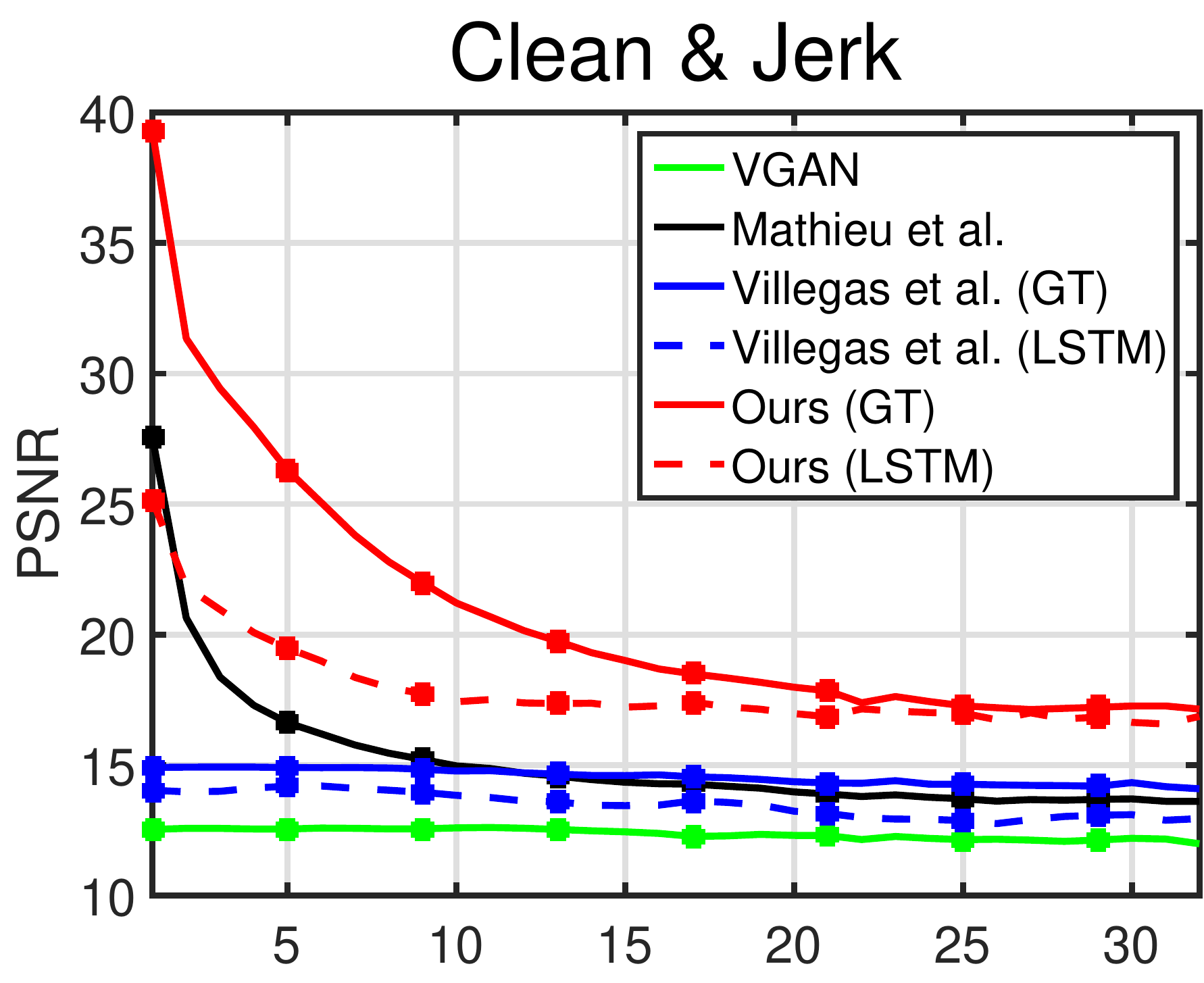}
\includegraphics[width=0.24\linewidth]{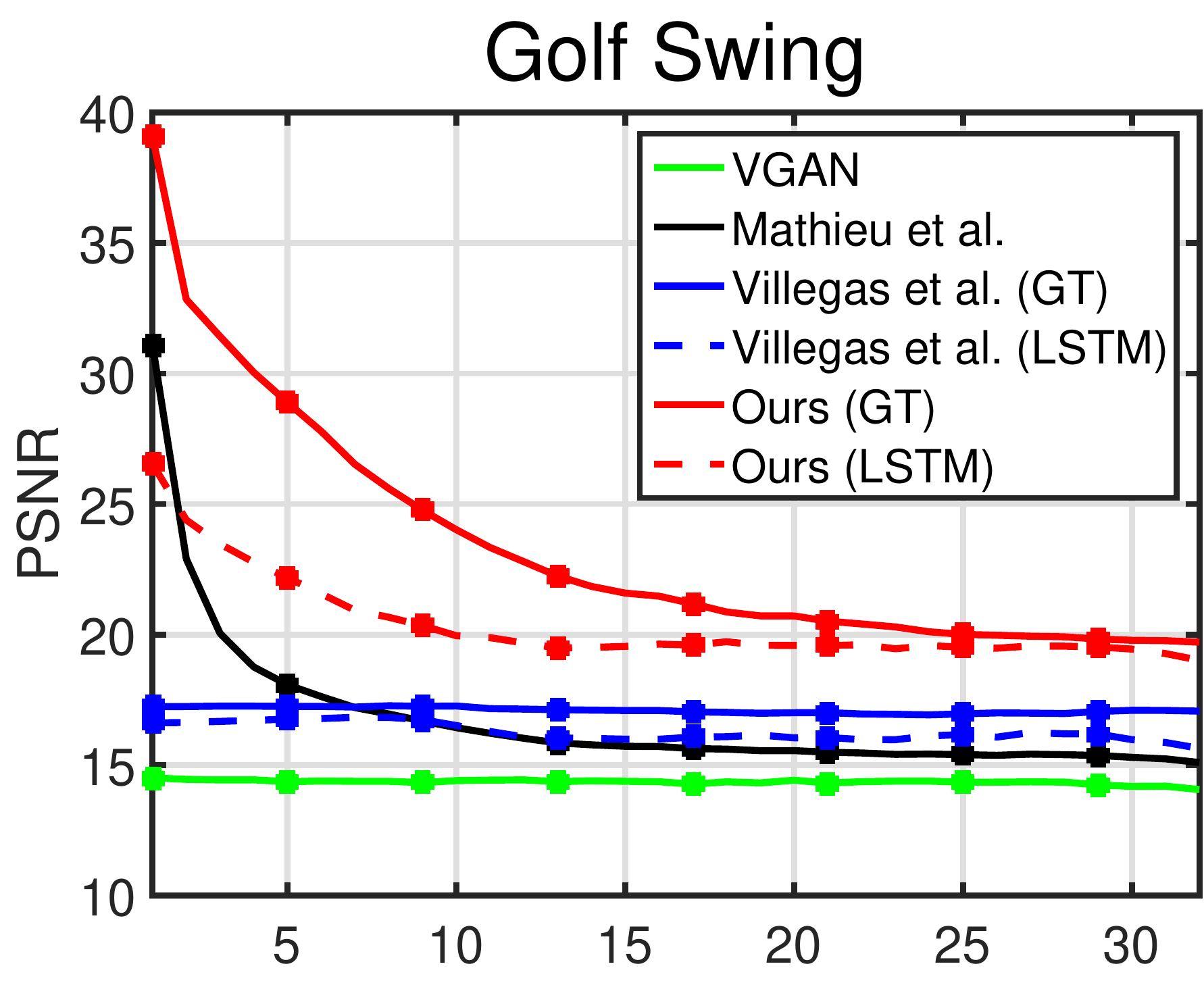}\\
\includegraphics[width=0.24\linewidth]{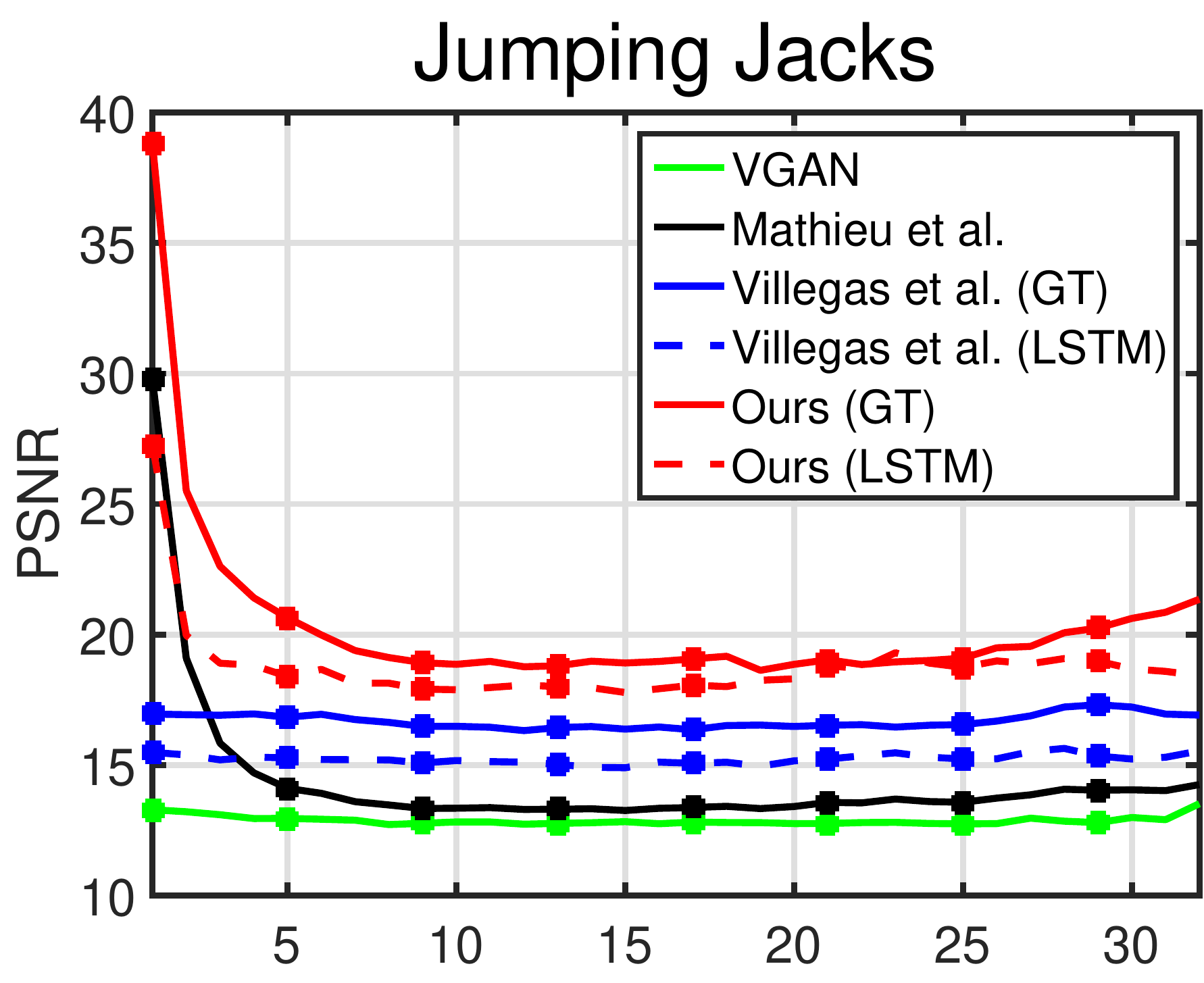}
\includegraphics[width=0.24\linewidth]{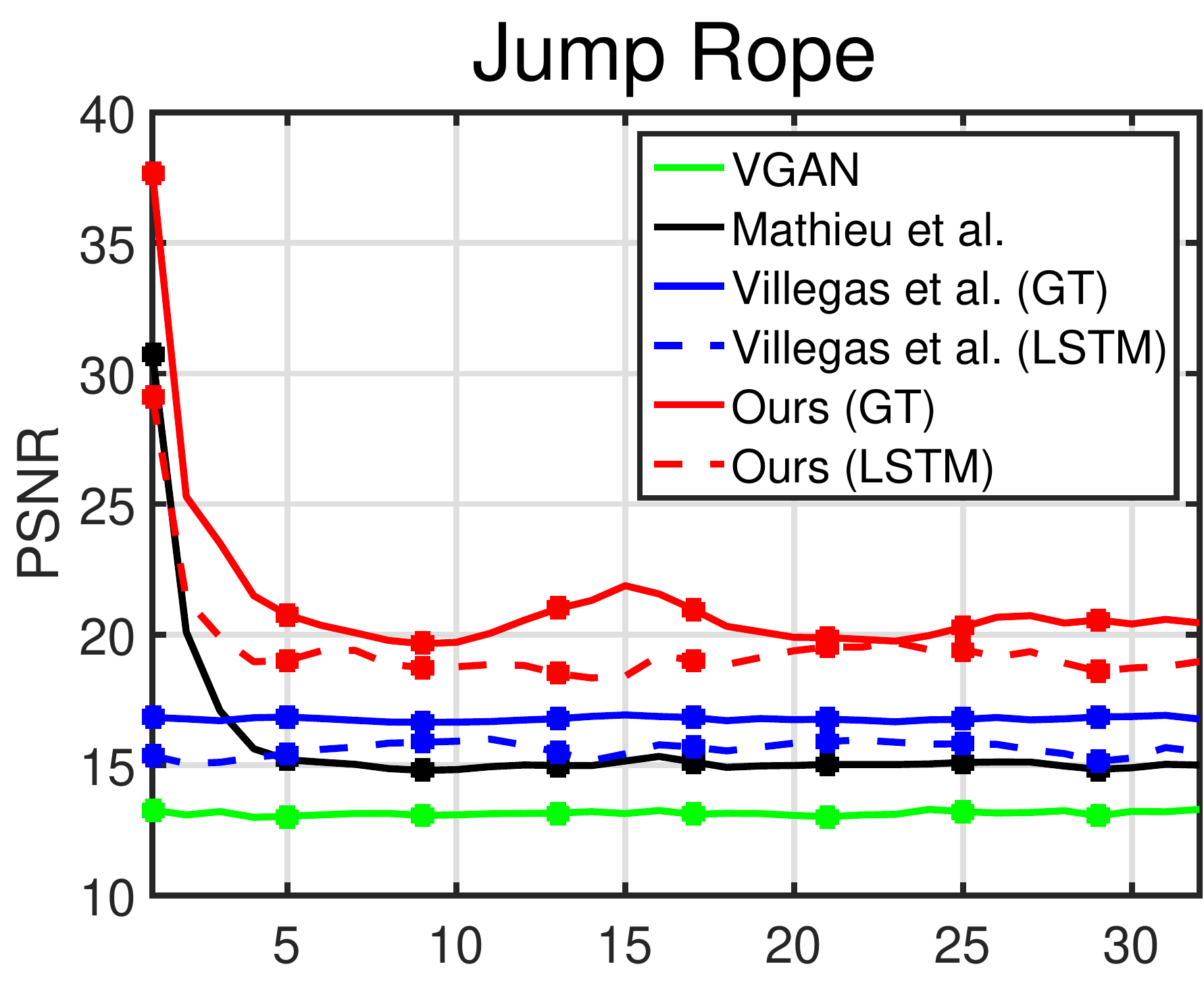}
\includegraphics[width=0.24\linewidth]{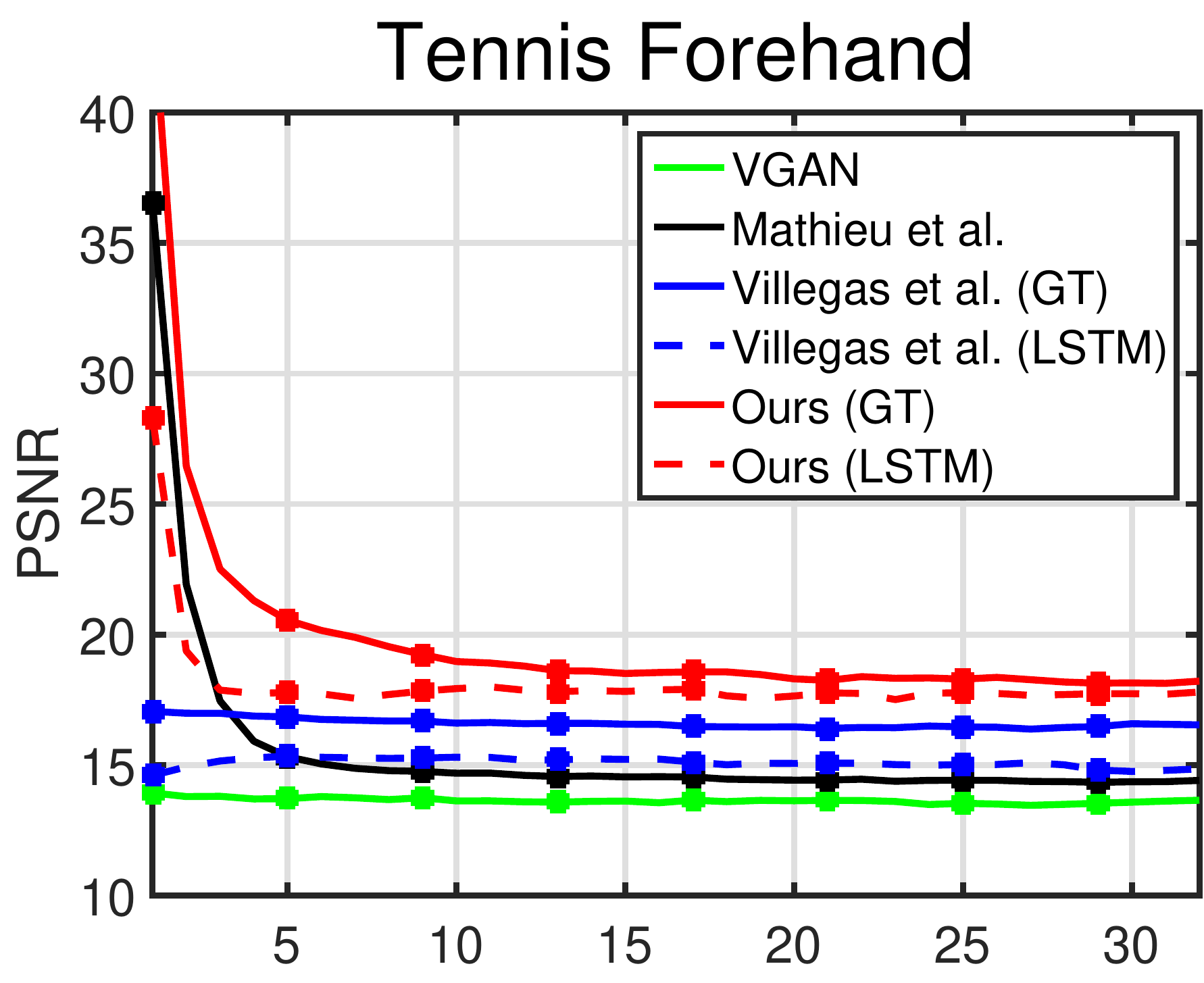}
\includegraphics[width=0.24\linewidth]{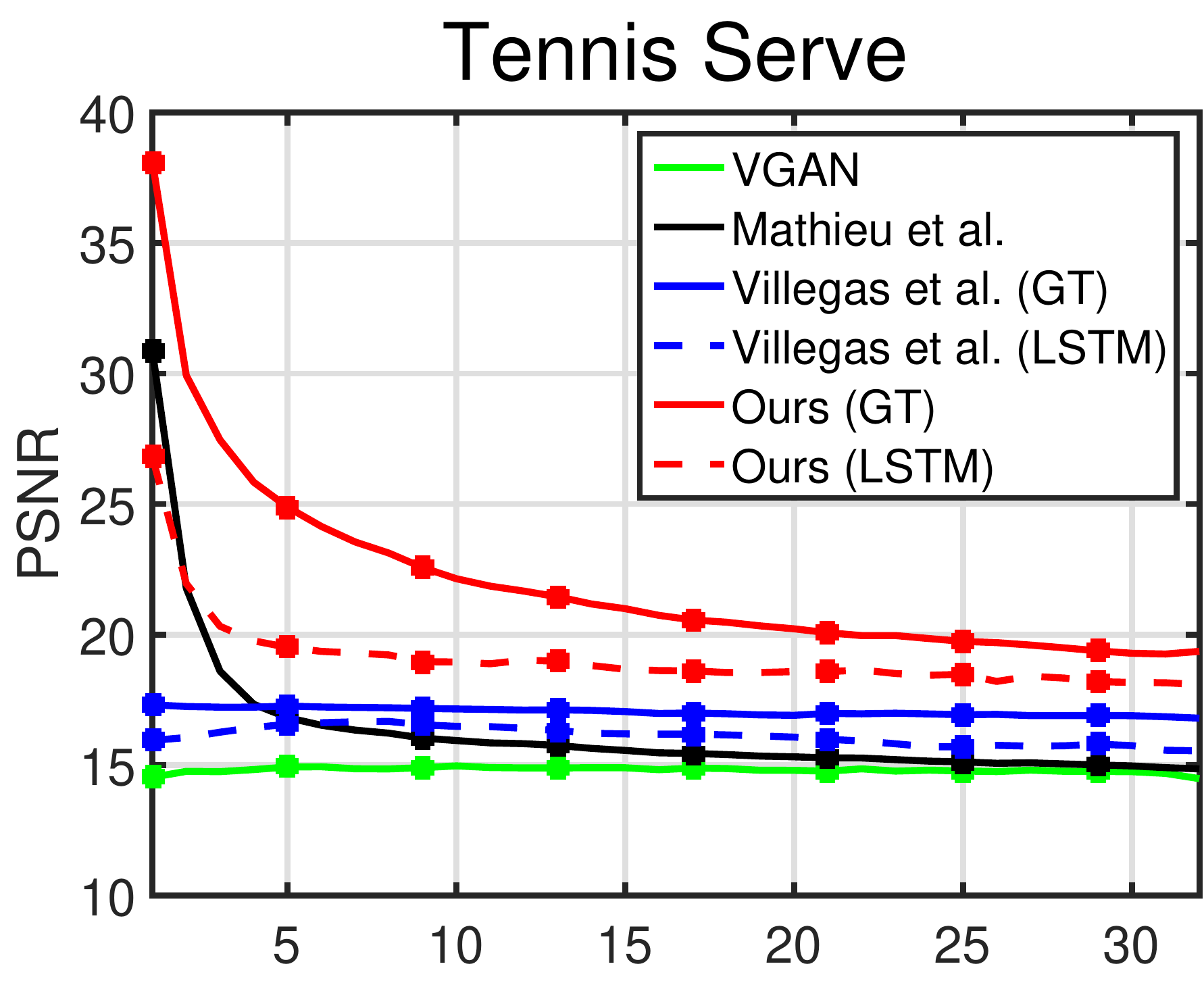}
\caption{Comparison of state-of-the-arts using Peak Signal-to-Noise Ratio (PSNR) on different human action categories from the Penn Action Dataset~\cite{zhang2013from}.}
\label{fig:PSNR}
\end{figure}

\p{Results.} We evaluate the predictions using Peak Signal-to-Noise Ratio (PSNR) and Mean Square Error (MSE). Both metrics perform pixel-level analysis between the ground truth frames and the generated videos. We also report the results of our method and~\cite{villegas2017learning} using the condition motion maps computed from the ground truth joints (GT). The results are shown in Figure~\ref{fig:PSNR} and Table~\ref{tbl:MSE} respectively. From these two scores, we discover that the proposed method achieves better quantitative results which demonstrates the effectiveness of our algorithm.

\begin{figure}[t]
\centering
\includegraphics[width=\linewidth]{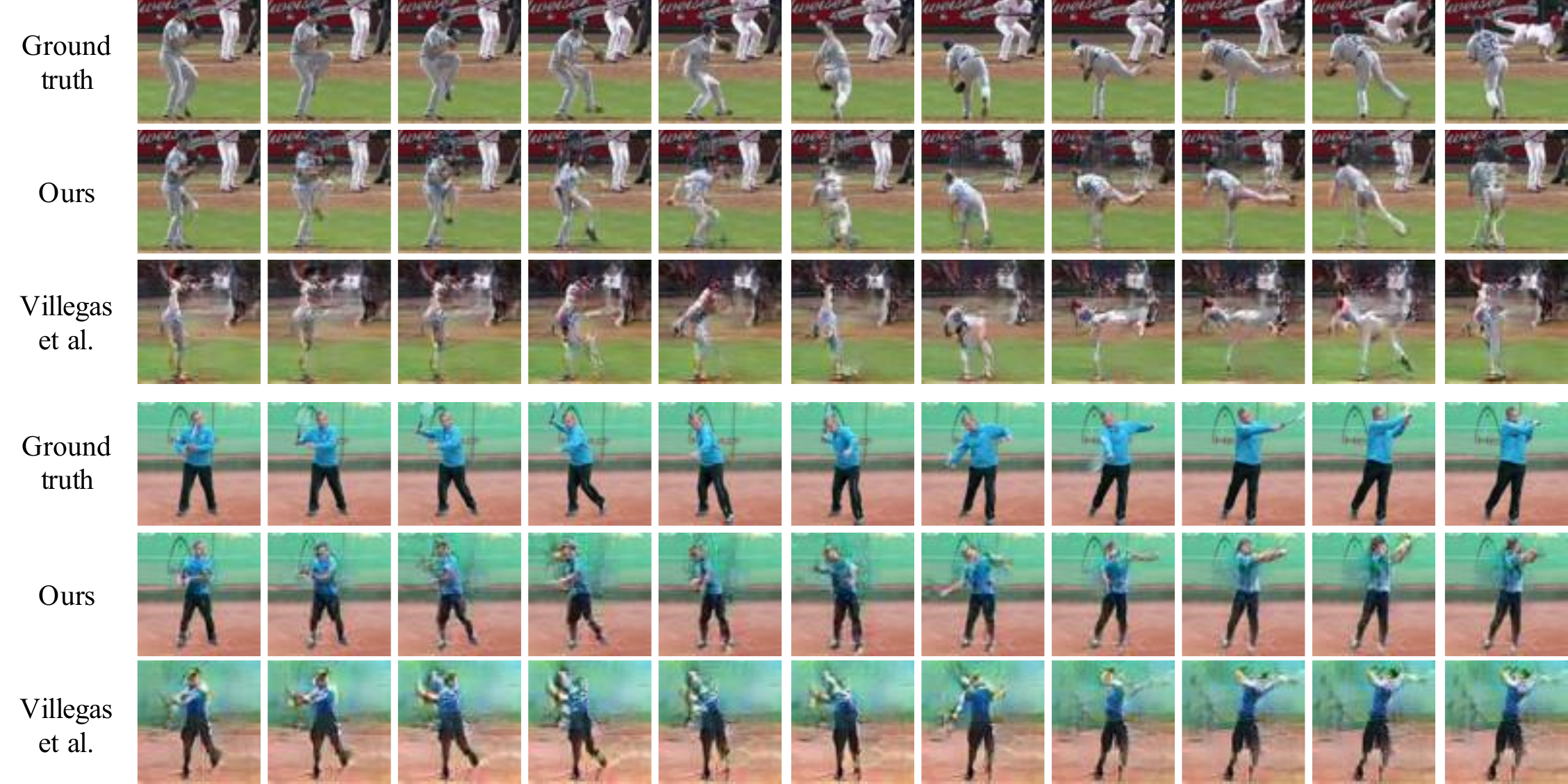}
\caption{Visual comparison of our method with Villegas et al.~\cite{villegas2017learning} on the Penn Action Dataset~\cite{zhang2013from}. Examples are sampled from the action of baseball (\textit{top}) and tennis (\textit{bottom}) respectively. The results are taken every 5 frames.}
\label{fig:PA}
\end{figure}

Figure~\ref{fig:PA} shows visual comparison of our method with~\cite{villegas2017learning}. We can find that the predicted future of our method is closer to the ground-true future. To be speclfic, our method yields more consistent motions and keeps human appearances as well. Due to space constraints, we ask the reader to refer to the project website for more side by side visual results.

\setlength{\tabcolsep}{2pt}
\begin{table}[t]
\begin{minipage}[t]{.46\linewidth}
\begin{center}
\caption{Comparison of state-of-the-art algorithms on the Penn Action Database~\cite{zhang2013from}. A smaller MSE score means better performance.}
\label{tbl:MSE}
\begin{tabular}{c|c|c}
\hline
Methods & MSE & MSE (GT)\\
\hline\hline
VGAN~\cite{vondrick2016generating} & 0.047 & -\\ \hline
Mathieu et al.~\cite{mathieu2016deep} & 0.041 & -\\ \hline
Villegas et al.~\cite{villegas2017learning} & 0.030 & 0.025\\ \hline
Ours & \textbf{0.023} & \textbf{0.011}\\
\hline
\end{tabular}
\end{center}
\end{minipage}
\hfill
\begin{minipage}[t]{.50\linewidth}
\begin{center}
\caption{Quantitative results of ablation study. We report the ACD-* scores on the MUG Database~\cite{aifanti2010mug} and MSE scores on the Penn Action Dataset~\cite{zhang2013from}.}
\label{tbl:ablation}
\begin{tabular}{c|c|c|c}
\hline
Settings & ACD-I & ACD-C & MSE\\
\hline\hline
$G_M$ \st{(Dense)}, \st{$G_R$} & 0.459 & 0.155 & 0.027\\ \hline
$G_M$ (Dense), \st{$G_R$} & 0.252 & 0.140 & 0.014\\ \hline
$G_M$ (Dense), $G_R$ & \textbf{0.184} & \textbf{0.107} & \textbf{0.011}\\
\hline
\end{tabular}
\end{center}
\end{minipage}
\end{table}
\setlength{\tabcolsep}{1.4pt}

\subsection{Ablation Study}
\label{sect:experiments:ablation}

Our method consists of three main modules: residual learning, dense connections for the decoder and the two-stage generation schema. Without residual learning, our network decays to~\cite{villegas2017learning}. As shown in Section~\ref{sect:experiments:expression} and~\ref{sect:experiments:pose}, ours outperforms~\cite{villegas2017learning} which demonstrates the effectiveness of residual learning. To verify the rest modules, we train one partial variant of $G_M$, where the dense connections are not employed in the decoder $f_D$. Then we evaluate three different settings of our method on both tasks: $G_M$ without dense connections, using only $G_M$ for generation and our full model. Note that in order to get rid of the influence from the LSTM, we report the results using the conditional motion maps calculated from the ground truth on the Penn Action Dataset. Results are shown in Table~\ref{tbl:ablation}. Our approach with more modules performs better than those with less components, which suggests the effectiveness of each part of our algorithm.

\section{Conclusions}
\label{sect:conclusions}

In this paper, we combine the benefits of high-level structural conditions and spatiotemporal generative networks for image-to-video translation by synthesizing videos in a generation and then refinement manner. We have applied this method to facial expression retargeting where we show that our method is able to engage domain knowledge for realistic video generation, and to human pose forecasting where we demonstrate that our method achieves higher performance than state-of-the-arts when generating videos involving large motion changes. We also incorporate residual learning and dense connections to produce high-quality results. In the future, we plan to further explore the use of our framework for other image or video generation tasks.

\vspace{3mm}
\noindent {\bf Acknowledgment}. This work is partly supported by the Air Force Office of Scientific Research (AFOSR) under the Dynamic Data-Driven Application Systems Program, NSF 1763523, 1747778, 1733843 and 1703883 Awards. Mubbasir Kapadia has been funded in part by NSF IIS-1703883, NSF S\&AS-1723869, and DARPA SocialSim-W911NF-17-C-0098.

%
%
%
\bibliographystyle{splncs04}
\bibliography{egbib}
\end{document}